\documentclass[conference]{IEEEtran}
\IEEEoverridecommandlockouts
\usepackage{cite}
\usepackage{amsmath,amssymb,amsfonts}
\usepackage{algorithm}
\usepackage{algorithmic}
\usepackage{graphicx}
\usepackage{textcomp}
\usepackage{xcolor}
\def\BibTeX{{\rm B\kern-.05em{\sc i\kern-.025em b}\kern-.08em
    T\kern-.1667em\lower.7ex\hbox{E}\kern-.125emX}}
    
\usepackage{microtype}
\usepackage{graphicx}
\usepackage{subfigure}
\usepackage{booktabs} 
\usepackage{times}  
\usepackage{helvet} 
\usepackage{courier}  
\usepackage[hyphens]{url}  
\usepackage{graphicx} 
\usepackage{caption} 

\usepackage{amsmath}
\usepackage{amssymb}
\usepackage{algorithm}
\usepackage{algorithmic}
\usepackage{color}
\usepackage{tikz}
\usepackage{pgfplots}
\pgfplotsset{compat=1.16}
\usepackage{pgffor}

\usepackage{listings}

\begin{document}

\title{Improving Model and Search for Computer Go\\
\thanks{
This work was granted access to the HPC resources of IDRIS under the allocation 2020-AD011012119 made by GENCI. 

This work was supported in part by the French government under management of Agence Nationale de la Recherche as part of the “Investissements d’avenir” program, reference ANR19-P3IA-0001 (PRAIRIE 3IA Institute).}
}

\author{\IEEEauthorblockN{Tristan Cazenave}
\IEEEauthorblockA{\textit{LAMSADE} \\
\textit{Universit\'e Paris-Dauphine, PSL, CNRS}\\
\textit{Paris, France}\\
Tristan.Cazenave@dauphine.psl.eu}
}

\maketitle

\begin{abstract}
The standard for Deep Reinforcement Learning in games, following Alpha Zero, is to use residual networks and to increase the depth of the network to get better results. We propose to improve mobile networks as an alternative to residual networks and experimentally show the playing strength of the networks according to both their width and their depth. We also propose a generalization of the PUCT search algorithm that improves on PUCT.
\end{abstract}

\begin{IEEEkeywords}
Monte Carlo Tree Search, Deep Learning, Computer Games
\end{IEEEkeywords}

\section{Introduction}

Training deep neural networks and performing tree search are the two pillars of current board games programs. Deep reinforcement learning combining self play and Monte Carlo Tree Search (MCTS) \cite{Coulom2006,Kocsis2006} with the PUCT algorithm \cite{silver2016mastering} is the current state of the art for computer Go \cite{silver2017mastering,greatfast} and for other board games \cite{silver2018general,cazenave2020polygames}

MobileNets have already been used in computer Go \cite{cazenave2020mobile}. In this paper we improve on this approach evaluating the benefits of improved MobileNets for computer Go using Squeeze and Excitation in the inverted residuals and using a multiplication factor of 6 for the planes of the trunk. We get large improvements both for the accuracy and the value when previous work obtained large improvements only for the value.

Some papers only take the accuracy of networks and the number of parameters into account. For games the speed of the networks is a critical property since the networks are used in a real time search engine. There is a dilemma between the accuracy and the speed of the networks. We experiment with various depth and width of network and find that when increasing the size of the networks there is a balance to keep between the depth and the width of the networks.

We are also interested in improving the Monte Carlo Tree Search that uses the trained networks. We propose Generalized PUCT (GPUCT), a generalization of PUCT that makes the best constant invariant to the number of descents.












The remainder of the paper is organized as follows. The second section presents related works in Deep Reinforcement Learning for games. The third section describes the Generalized PUCT bandit. The fourth section details the neural networks we trained. The fifth section gives experimental results.

\section{Previous Work}

\subsection{Zero Learning}

Monte Carlo Tree Search (MCTS) \cite{Coulom2006,Kocsis2006} made a revolution in Artificial Intelligence applied to board games. A second revolution occurred when it was combined with Deep Reinforcement Learning which led to superhuman level of play in the game of Go with the AlphaGo program \cite{silver2016mastering}.

Residual networks \cite{Cazenave2018residual}, combined with policy and value heads sharing the same network and Expert Iteration \cite{anthony2017thinking} did improve much on AlphaGo leading to AlphaGo Zero \cite{silver2017mastering} and zero learning. With these improvements AlphaGo Zero was able to learn the game of Go from scratch and surpassed AlphaGo.

There were many replication of AlphaGo Zero, both for Go and for other games. For example ELF/OpenGo \cite{tian2019elfopengo}, Leela Zero \cite{pascutto2017leela}, Crazy Zero by Coulom and the current best open source Go program KataGo \cite{greatfast}.

The approach was also used for learning a large set of games from zero knowledge with Galvanise Zero \cite{gzero} and Polygames \cite{cazenave2020polygames}.

\subsection{Neural Architectures}

AlphaGo used a convolutional network with 13 layers and 256 planes.

Current computer Go and computer games programs use a neural networks with two heads, one for the policy and one for the value as in AlphaGo Zero  \cite{silver2017mastering}.  Using a network with two heads instead of two networks was reported to bring a 600 ELO improvement and using residual networks \cite{Cazenave2018residual} also brought another 600 ELO improvement. The standard for Go programs is to follow AlphaGo Zero and use a residual netwrork with 20 or 40 blocks and 256 planes.

An innovation in the Katago program is to use Global Average Pooling in the network in some layers of the network combined with the residual layers. It also uses more than two heads as it helps regularization.

Polygames also uses Global Average Pooling in the value head. Together with a fully convolutional policy, it make Polygames networks invariant to changes in the size of the board.

\subsection{Mobile Networks}

MobileNet \cite{howard2017mobilenets} and then MobileNetV2 \cite{sandler2018mobilenetv2} are a parameter efficient neural network architecture for computer vision. Instead of usual convolutional layers in the block they use depthwise convolutions. The also use 1x1 filters to pass for a small number of channels in the trunk to 6 times more channels in the block. 

MobileNets were successfully applied to the game of Go \cite{cazenave2020mobile}. Our approach is a large improvement on this approach using Squeeze and Excitation and a 6 fold increase in the number of channels in the blocks. We also compare networks of different width and depth and show some possible choices for increasing width and depth are dominated and that it is better to increase both width and depth when making the networks grow.

\section{Improving the Search}

\subsection{PUCT}

The Monte Carlo Tree Search algorithm used in current computer Go programs since AlphaGo is PUCT. A Bandit that include the policy prior in its exploration term. The bandit for PUCT is:

$$V(s, a)= Q(s,a) + c \times P(s,a) \times \frac{\sqrt{N(s)}}{1+N(s,a)}$$

where $P(s,a)$ is the probability of move $a$ to be the best moves in state $s$ given by the policy head, $N(s)$ is the total number of descents performed in state $s$ and $N(s,a)$ is the number of descents for move $a$ in state $s$.

\subsection{Generalized PUCT}

We propose to generalize PUCT replacing the square root with an exponential and using a parameter $\tau$ for the exponential. The Generalized PUCT bandit (GPUCT) is:

$$V(s, a)= Q(s,a) + c \times P(s,a) \times \frac{e^{\tau \times log(N(s))}}{1+N(s,a)}$$

This is a generalization of PUCT since for $\tau=0.5$ this is the PUCT bandit.

\subsection{Experimental Results}

We experimented with various constants and numbers of evaluations for the PUCT bandit. We found that for small numbers of evaluations the 0.1 constant was performing well. In order to make the experiments stable we made PUCT with numbers of evaluations starting at 16 and doubling until 512 and a constant of 0.1 play 400 games against PUCT with the same number of evaluations but varying constants. The result are given in Table \ref{tableConstants}. The last line of the table is the average winning rate. On average the 0.15 constant is the best and the following constants have decreasing average.

\begin{table}[h]
  \centering
  \caption{Evolution of win rates with the constant and the number of descents. Average over 400 games against the 0.1 constant. $\frac{\sigma}{\sqrt{n}} \in (2.0 ; 2.5)$}
  \label{tableConstants}
  \begin{tabular}{lrrrrrrrrrrrrrrrrrr}
$d ~ \backslash ~ c$ &  0.05 &  0.15 &  0.20 &  0.25 &  0.30 &   0.35 \\
16                   & 57.75 & 50.25 & 47.25 & 46.50 & 43.50 &  44.00 \\
32                   & 48.75 & 54.00 & 48.50 & 48.75 & 52.50 &  49.50 \\
64                   & 45.25 & 56.75 & 55.00 & 50.25 & 47.75 &  44.25 \\
128                  & 33.50 & 49.00 & 50.25 & 47.50 & 48.00 &  47.25 \\
256                  & 34.75 & 50.25 & 56.75 & 51.50 & 45.00 &  44.75 \\
512                  & 35.50 & 59.50 & 56.75 & 66.00 & 61.25 &  60.25 \\
\hline
avg                  & 42.58 & 53.29 & 52.42 & 51.75 & 49.67 & 48.33 \\
  \end{tabular}
\end{table}

It is clear from this table that there is a drift of the best constant towards greater values. We repair this undesirable property with Generalized PUCT. In order to find the best $\tau$ for Generalized PUCT we used the following algorithm:

$$argmin _{\tau,c} (\Sigma_d |c \times e^{\tau \times log(d)} - c_d \times e^{0.5 \times log(d)}|)$$

where $d$ is the budget (the first column of Table \ref{tableConstants}), $c_d$ is the best PUCT constant for budget $d$ (for example 0.15 for $d = 32$).

Given the data in Table \ref{tableConstants} the best parameters values we found are $\tau=0.737$ and $c=0.057$.

In order to verify that these values counter the drift in the best constants of PUCT, we made PUCT with different budgets and constants play against Generalized PUCT with $\tau=0.737$ and $c=0.057$ and the same budget as PUCT. The results are given in Table \ref{tableGPUCT}. For small budgets the PUCT algorithm with a constant of 0.2 is close to GPUCT but when the budget grows to 1024 or 2048 descents, GPUCT gets much better than PUCT. This is also the case for PUCT with the 0.1 and a 0.15 constants.

\begin{table}[h]
  \centering
  \caption{Comparison of PUCT with constants of 0.1, 0.15 and 0.2 against GPUCT with $\tau=0.737$ and $c=0.057$. 400 games. $\frac{\sigma}{\sqrt{n}} \in (2.0 ; 2.5)$}
  \label{tableGPUCT}
  \begin{tabular}{lrrrrrrrrrrrrrrrrrr}
$d$ &  ~~~~~~~~~~~0.1 & ~~~~~~~~~~~0.15 & ~~~~~~~~~~~0.2 \\
16                   & 57.75 & 51.50 & 61.25 \\
32                   & 57.00 & 48.50 & 54.75 \\
64                   & 56.00 & 45.25 & 48.50 \\
128                  & 48.75 & 49.50 & 45.00 \\
256                  & 59.50 & 55.00 & 55.00 \\
512                  & 55.00 & 61.25 & 53.00 \\
1024                 & 73.75 & 63.00 & 53.75 \\
2048                 &       & 67.00 & 65.50 \\
  \end{tabular}
\end{table}

Our findings are consistent with the PUCT constant used in other zero Go programs. For example ELF uses a 1.5 constant \cite{tian2019elfopengo} for much more tree descents than we do. Our model fits this increase of the value of the best constant when the search uses more descents.

We did not try to tune the constants for GPUCT and still get better results than PUCT. Some additional strength may come from tuning the constants instead of only using the two constants coming from optimizing on the data of Table \ref{tableConstants}.


Note that having the same constant for different budgets can also be useful for programs that use pondering or adaptive thinking times. Besides being more convenient for tuning the constant.

\section{Improving the Model}

In this section we start describing the dataset we built. We then detail the training and the inputs and labels. We also explain how we have added Squeeze and Excitation to the MobilNets and how we experimented with various depth and width of the networks. We finally give experimental results for various networks.

\subsection{The Katago dataset}

Katago is the strongest available computer Go program. It has released the self-played games of 2020 as sgf files. We selected from these games the games played on a 19x19 board with a komi between 5.5 and 7.5. We built the Katago dataset taking the last 1,000,000 games played by Katago. The validation dataset is bult by randomly selecting 100,000 games from the 1,000,000 games and taking one random state from each game. The games of the validation set are never used during training.

The Katago dataset is a better dataset than the ELF and the Leela datasets used in \cite{cazenave2020mobile}. Katago plays Go at a much better level than ELF and Leela. The networks trained on the Katago dataset are trained with better data.

\subsection{Training}

Networks are trained with Keras with each epoch containing 1 000 000 states randomly taken in the Katago dataset.

The evaluation on the validation set is computed every epoch. The loss used for evaluating the value is the mean squared error (MSE) as in AlphaGo. However we train the value with the binary cross entropy loss. The loss for the policy is the categorical cross entropy loss. We evaluate the policy according to its accuracy on the validation set. 

The batch size is 32 due to memory constraints. The annealing schedule is to train with a learning rate of 0.0005 for the first 100 epochs, then to train with a learning rate of 0.00005 for 50 epochs, and then with 0.000005 for 50 epochs, and finally with 0.0000005 for another 50 epochs. An epoch takes between 3 minutes for the smallest network and 30 minutes for the larger ones using a V100 card.

All layers in the networks have L2 regularization with a weight of 0.0001. The loss is the sum of the value loss, the policy loss and the L2 loss.

\subsection{Inputs and Labels}

The inputs of the networks use the colors of the stones, the liberties, the ladder status of the stone, the ladder status of adjacent strings (i.e. if an adjacent string is in ladder), the last 5 states and a plane for the color to play. The total number of planes used to encode a state is 21.

The labels are a 0 or a 1 for the value head. A 0 means Black has won the game and a 1 means it is White. For the policy head there is a 1 for the move played in the state and 0 for all other moves.

\subsection{Squeeze and Excitation}

We add Squeeze and Excitation \cite{hu2018squeeze} to the MobileNets so as to improve their performance in computer Go. The way we do it is by adding a Squeeze and Excitation block at the end of the MobileNet block before the addition. 

The squeeze and excitation block starts with Global Average Pooling followed by two dense layers and a multiplication of the input tensor by the output of the dense layers.

We give here the Keras code we used for this block:

\begin{lstlisting}[language=Python]
def SE_Block(t,filters,ratio=16):
    se_shape = (1, 1, filters)
    se = GlobalAveragePooling2D()(t)
    se = Reshape(se_shape)(se)
    se = Dense(filters // ratio,
               activation='relu',
               use_bias=False)(se)
    se = Dense(filters,
               activation='sigmoid',
               use_bias=False)(se)
    x = multiply([t,se])
    return x
\end{lstlisting}

\subsection{Depth and Width of Mobile Networks}

In order to improve the performance of AlphaGo Zero their authors made the network grow from 20 residual blocks of 256 planes to 40 residual blocks of 256 planes. 

In this paper we claim that MobileNets with Squeeze and Excitation give much better results that residual networks for the game of Go and we also claim that to improve the performance of a network it is better to make both the number of blocks (i.e. the depth of the network) and the number of planes (i.e. the width of the network) grow together. 

\section{Experimental Results}

\subsection{Networks with Less than One Million Parameters}

All the networks we test use two heads, one for the policy and one for the value. The Alpha Zero like network uses fully connected layers both in the policy and the value head as in AlphaGo and descendants. When restricted to one million parameters it is detrimental since it uses a little less than 300,000 parameters only for the two heads. The remaining of the AlphaZero network is 10 blocks of 44 planes. For all the networks we optimize the number of parameters so as to be as close to one million paramaters as possible. The Polygames network use a fully convolutional policy head, with no parameters in the policy head. It also uses Global Average Pooling in the value head before a fully connected layer of 50 outputs and then a fully connected layer whit one output for the value. The Polygames network uses 13 residual blocks of 64 planes. The Mobile and the Mobile+SE networks use 17 mobile blocks of 384 planes with a trunk of 64 planes and the Polygames heads.

\begin{figure}
\centering
\includegraphics[scale=0.5]{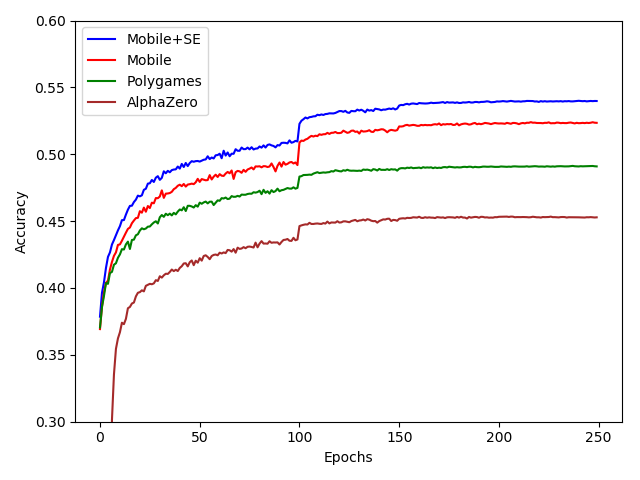}
\caption{The evolution of the policy validation accuracy for the different networks with less than one million parameters on the Katago dataset.}
\label{smallPolicy}
\end{figure}

The evolution of the policy accuracy for the four networks is given in Figure \ref{smallPolicy}. Polygames architecture gives better results than the Alpha Zero architecture. Mobile is better than Polygames and Mobile+SE is better than Mobile alone.

\begin{figure}
\centering
\includegraphics[scale=0.5]{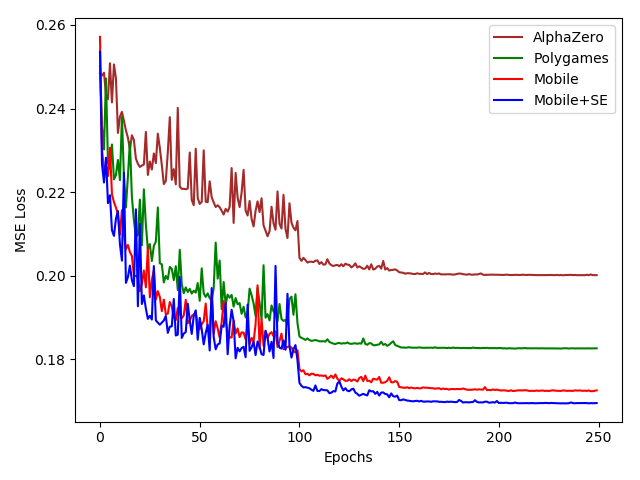}
\caption{The evolution of the value validation MSE loss for the different networks with less than one million parameters on the Katago dataset.}
\label{smallValue}
\end{figure}

Figure \ref{smallValue} gives the evolution of the MSE validation loss of the value during training. Again Mobile+SE is better than Mobile. Mobile is better than Polygames and Polygames is better than AlphaZero.

\subsection{Training Large Networks}

Multiple MobileNets and two Polygames/Alpha Zero like residual networks were trained on the Katago dataset. It took a total of more than 10,000 hours of training using V100 cards.

We trained a 20 blocks and a 40 blocks residual network with the Polygames heads. The results for these two networks are given in Table \ref{tableResidual}.

We trained many MobileNets with Squeeze and Excitation and the Polygames heads. The number of parameters of the networks according to their width and depth are given in Table \ref{tableParameters}. The GPU speed of the networks are given in Table \ref{tableSpeed}. The CPU speed of the networks are given in Table \ref{tableSpeedCPU}. The accuracy reached on the validation set at the end of training are given in Table \ref{tableAccuracy}. The MSE validation loss of the value is given in Table \ref{tableMSE}.

\begin{table*}
  \centering
  \caption{Properties of residual networks.}
  \label{tableResidual}
  \begin{tabular}{lrrrrrrrrrr}
  
  Network                                    &  ~~~~~Parameters & Accuracy & ~~~~~~MSE & GPU Speed with Batch=32 & CPU Speed with Batch=1\\
                                             &        &         & \\
residual.20.256 & 23,642,469 & 55.12 & 0.1667 & 21.30 & 6.04\\
residual.40.256 & 47,266,149 & 55.21 & 0.1680 & 13.54 & 3.42\\
  \end{tabular}
\end{table*}

\begin{table*}[h]
  \centering
  \caption{Parameters according to width and depth.}
  \label{tableParameters}
  \begin{tabular}{lrrrrrrrrrrrrrrrrrr}
$w ~ \backslash ~ d$ &         16 &         32 &         48 &         64 &        80\\
64                   &    908,197 &  1,811,365 &  2,714,533 &  3,617,701 & 4,520,869\\
96                   &  1,958,213 &  3,908,933 &  5,859,653 &  7,810,373 & 9,761,093 \\
128                  &  3,405,541 &  6,801,125 & 10,196,709 & 13,592,293 & \\
160                  &  5,250,181 & 10,487,941 & 15,725,701 & 20,696,901 & \\
192                  &  7,492,133 & 14,969,381 & 22,446,629 & 29,923,877 & \\
224                  & 10,131,397 & 20,245,445 &            &            & \\
  \end{tabular}
\end{table*}

\begin{table}[h]
  \centering
  \caption{Speed according to width and depth. Number of batches of size 32 per second on GPU (RTX 2080 Ti).}
  \label{tableSpeed}
  \begin{tabular}{lrrrrrrrrrrrrrrrrrr}
$w ~ \backslash ~ d$ &    16 &    32 &    48 &    64 &    80 &    96 \\
64                   & 28.17 & 25.28 & 19.39 & 18.30 & 16.20 & 14.13 \\
96                   & 26.61 & 21.65 & 16.56 & 13.51 & 11.62 &  9.90 \\
128                  & 25.25 & 17.65 & 13.32 & 10.71 &  9.07 &  7.71 \\
160                  & 21.82 & 14.37 & 10.44 &  8.18 &  6.87 &  5.94 \\
192                  & 19.78 & 12.33 &  9.02 &  6.97 &  5.60 &  4.81 \\
224                  & 15.93 & 10.33 &  7.39 &  5.69 &  4.66 &  3.83 \\
256                  & 14.08 &  8.15 &  6.26 &  4.81 &  3.95 &  3.32 \\
  \end{tabular}
\end{table}

\begin{table}[h]
  \centering
  \caption{Speed according to width and depth. Number of batches of size 1 per second on CPU.}
  \label{tableSpeedCPU}
  \begin{tabular}{lrrrrrrrrrrrrrrrrrr}
$w ~ \backslash ~ d$ &    16 &    32 &    48 &    64 &    80 \\
64                   & 25.53 & 15.57 & 12.79 &  9.83 &  7.93 \\
96                   & 19.80 & 11.57 &  9.39 &  7.54 &  5.91 \\
128                  & 14.48 &  8.85 &  6.89 &  5.49 & \\
160                  &  6.86 &  4.37 &  3.83 &       & \\
192                  &  5.87 &  3.61 &  2.52 &  1.88 & \\
224                  &  5.57 &  3.70 &       &       & \\
  \end{tabular}
\end{table}

\begin{table}[h]
  \centering
  \caption{Accuracy according to width and depth.}
  \label{tableAccuracy}
  \begin{tabular}{lrrrrrrrrrrrrrrrrrr}
$w ~ \backslash ~ d$ &    16 &    32 &    48 &    64 & 80\\
64                   & 53.98 & 55.94 & 56.98 & 57.77 & 58.21 \\
96                   & 55.48 & 57.78 & 58.51 & 59.20 & 59.62 \\
128                  & 56.52 & 58.56 & 59.40 & 60.06 &\\
160                  & 57.00 & 59.26 & 60.16 &       &\\
192                  & 57.65 & 59.73 & 60.97 & 61.28 &\\
224                  & 58.01 & 60.05 &       &       &\\
  \end{tabular}
\end{table}

\begin{figure}
\centering
\includegraphics[scale=0.5]{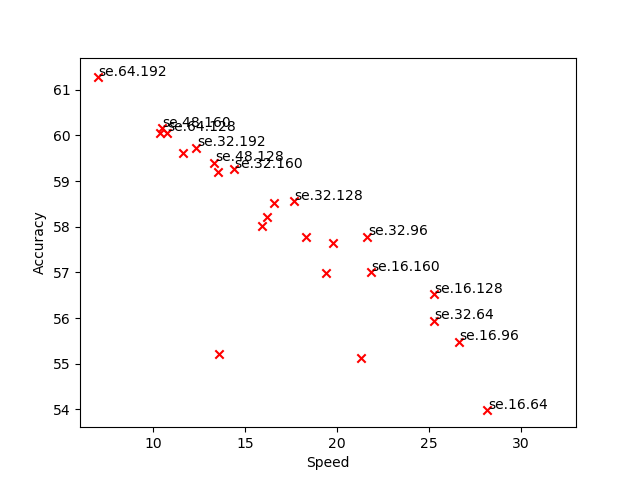}
\caption{The accuracy Pareto front.}
\label{Pareto.Accuracy}
\end{figure}

\begin{figure}
\centering
\includegraphics[scale=0.5]{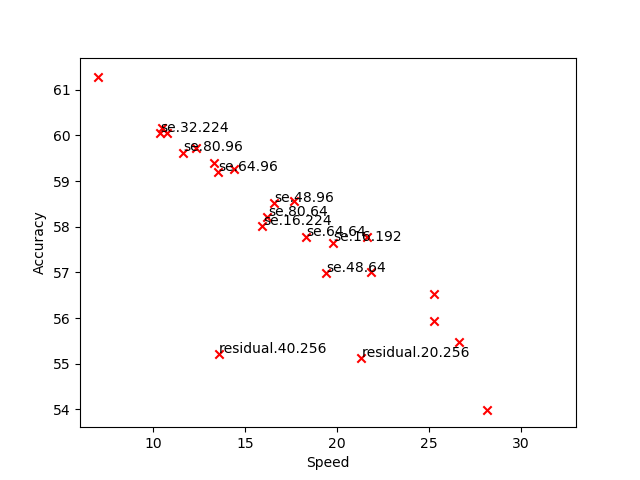}
\caption{Networks dominated by the accuracy Pareto front.}
\label{Pareto.Accuracy.Dominated}
\end{figure}

Figure \ref{Pareto.Accuracy} gives the Pareto front of the networks according to GPU speed and accuracy. Figure \ref{Pareto.Accuracy.Dominated} gives the networks that are dominated by other networks. The dominated networks are residual.20.256, residual.40.256, se.16.192 ,se.16.224, se.32.224, se.48.64, se.48.96, se.64.64, se.64.96, se.80.64, se.80.96. We can see in this list that networks that are either to shallow and too wide or to deep and too narrow are dominated. It means that there is a balance to keep between the depth and the width of the networks. Networks that are shallow and wide or networks that are deep and narrow are dominated. The optimal ratio $\frac{width}{depth}$ seems to lie somewhere between 2.67 and 6.00. But we have not enough data to assess whether it stays in the same range for greater depth and width. 

The accuracy of our MobileNets are much better than the accuracy previously reported for MobileNets that were close to the accuracy of residual nets for similar speeds \cite{cazenave2020mobile}. MobileNets that have the same GPU speed as the 20 blocks residual network have an accuracy close to 57\% whereas the residual net is close to 53\%. Moreover the Katago dataset has better quality games and is more elaborate than the datasets used in \cite{cazenave2020mobile}. It is even better when comparing the 40 blocks residual network to MobileNets with the same speed.

\begin{table}[h]
  \centering
  \caption{MSE according to width and depth.}
  \label{tableMSE}
  \begin{tabular}{lrrrrrrrrrrrrrrrrrr}
$w ~ \backslash ~ d$ &     16 &     32 &     48 &     64 &     80 \\
64                   & 0.1695 & 0.1637 & 0.1614 & 0.1602 & 0.1592 \\
96                   & 0.1657 & 0.1604 & 0.1580 & 0.1572 & 0.1563 \\
128                  & 0.1631 & 0.1583 & 0.1566 & 0.1553 &\\
160                  & 0.1615 & 0.1570 & 0.1551 &        &\\
192                  & 0.1603 & 0.1560 & 0.1536 & 0.1532 &\\
224                  & 0.1595 & 0.1558 &        &        &\\
  \end{tabular}
\end{table}

\begin{figure}
\centering
\includegraphics[scale=0.5]{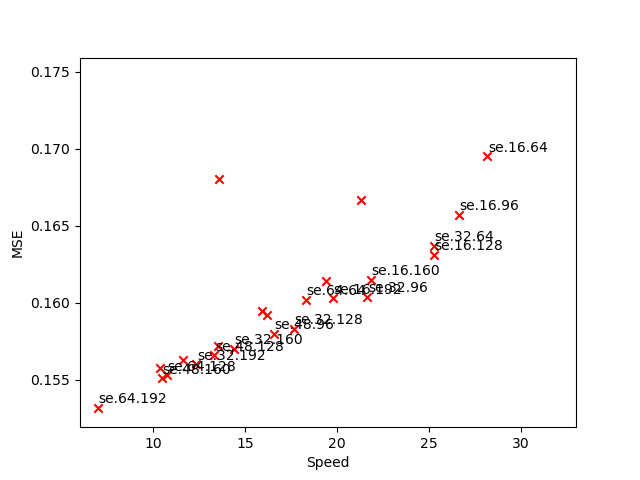}
\caption{The value Pareto front.}
\label{Pareto.Value}
\end{figure}

\begin{figure}
\centering
\includegraphics[scale=0.5]{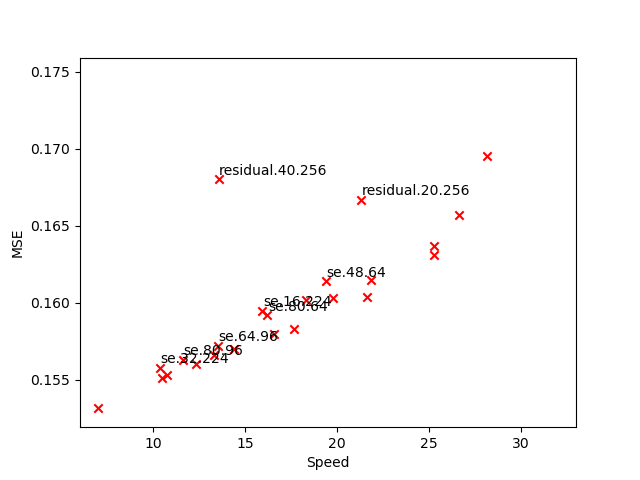}
\caption{Networks dominated by the value Pareto front.}
\label{Pareto.Value.Dominated}
\end{figure}

Figure \ref{Pareto.Value} gives the value Pareto front. Figure \ref{Pareto.Value.Dominated} gives the networks dominated by the value Pareto front. The dominated networks are residual.20.256, residual.40.256, se.16.224, se.32.224, se.48.64, se.64.96, se.80.64 and se.80.96. Again we see that shallow and wide networks as well as deep and narrow networks are dominated.

The residual networks are largely dominated by squeeze and excitation MobileNets both for the accuracy and for the value.

\subsection{Extrapolation of the Accuracy}

In order to extrapolate the accuracy of the bigger networks we made a regression on a formula to estimate the accuracy given the depth and the width of the networks. We assume that the increase in accuracy is logarithmic with the size of the network and we find the appropriate parameters using this algorithm:

$$argmin _{p,p_1,p_2,p_3,p_4} (\Sigma_{d,w} |p - \frac{p_1}{d} - \frac{p_2}{w} - \frac{1}{\frac{d}{p_3} + \frac{w}{p_4}} - A(d,w)|^2)$$

where $d$ is the depth, $w$ the width and $A(d,w)$ the value of the accuracy in Table \ref{tableAccuracy}. The best parameters values we found are $p=64.2$, $p_1=70.5$, $p_3=1290$ and $p_4=390$ giving a minimal error of 1.39 for the 23 accuracies.

Table \ref{tableExtrapolation} gives the Pareto front for the extrapolated accuracies and the accuracies we experimentally found. The extrapolated accuracies are in parenthesis. The non dominated accuracies according to the speed and accuracies of the other networks are in bold. We can observe that the trend of balancing the depth and the width of the networks continues for extrapolated values.

\begin{table}[h]
  \centering
  \caption{Extrapolation of the accuracy Pareto front.}
  \label{tableExtrapolation}
  \begin{tabular}{lrrrrrrrrrrrrrrrrrr}
$w ~ \backslash ~ d$ &    16 &    32 &    48 &    64 &     80 \\
64           & \bf 53.98 & \bf 55.94 & 56.98 & 57.77 & 58.21\\
96           & \bf 55.48 & \bf 57.78 & 58.51 & 59.2 & 59.62\\
128           & \bf 56.52 & \bf 58.56 & \bf 59.40 & \bf 60.06 & \bf (60.53)\\
160           & \bf 57.00 & \bf 59.26 & \bf 60.16 & (60.74) & (61.02)\\
192           & 57.65 & \bf 59.73 & \bf 60.97 & \bf 61.28 & (61.36)\\
224           & 58.01 & 60.05 &  (60.97) & \bf (61.37) & \bf (61.62)\\
256           & (58.18) & (60.41) & (61.18) & \bf (61.57) & \bf (61.81)\\
  \end{tabular}
\end{table}

\subsection{Evolution of the Accuracy and of the Value Loss During Training}

\begin{figure}
\centering
\includegraphics[scale=0.5]{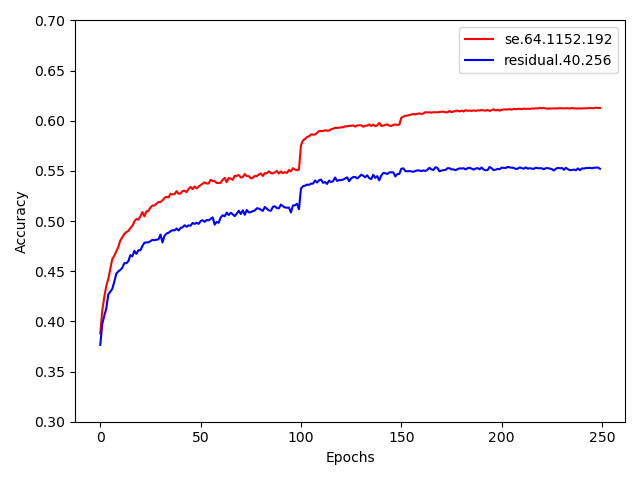}
\caption{The evolution of the policy validation accuracy for the best SE network and the 40 blocks residual network on the Katago dataset.}
\label{se.Accuracy}
\end{figure}

Figure \ref{se.Accuracy} gives the evolution of the accuracy of our best network against the evolution of the accuracy of a state of the art residual network. We can observe that training starts better for our network and that it improves more during training.

\begin{figure}
\centering
\includegraphics[scale=0.5]{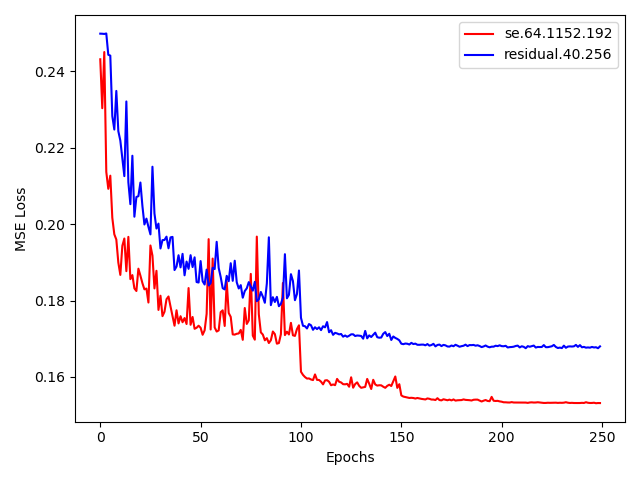}
\caption{The evolution of the value validation MSE loss for the best SE network and the 40 blocks residual network on the Katago dataset.}
\label{se.MSE}
\end{figure}

Figure \ref{se.MSE} gives the same evolution for the mean squared error of the value and again the residual network is largely dominated.

\subsection{Making the Networks Play}

We now make the networks play Go. We first test the strength of the networks only using the policy to play. Networks then plays instantly and still play at the level of high amateur Dan players. People enjoy playing blitz games on the internet against such networks. Table \ref{tablePolicies} gives the result of a round robin tournament between the policy networks. The name of a network is composed of the architecture ('se' for MobileNet with Squeeze and Excitation and 'residual' for residual networks) followed by the number of blocks, the number of planes in the inverted residual block and the number of planes in the trunk. For example se.64.1152.192 means a MobileNet with 64 blocks of 1152 planes and a trunk of 192 planes. For residual networks, residual.40.256 means a residual network of 40 blocks and 256 planes. As expected the se.64.1152.192 network, the one with the best accuracy has the best winning rate and the residual networks have results much worse than the best MobileNets.

\begin{table}
  \centering
  \caption{Round robin tournament between policies.}
  \label{tablePolicies}
  \begin{tabular}{lrrrrrrrrrr}
  
  Network                                    &  Winrate & ~~~~~~~~~$\frac{\sigma}{\sqrt{n}}$\\
                                             &        &         & \\
se.64.1152.192  & 0.833 & 0.054\\
se.48.1152.192  & 0.812 & 0.056\\
se.48.960.160   & 0.792 & 0.059\\
se.64.576.96    & 0.729 & 0.064\\
se.32.1344.224  & 0.688 & 0.067\\
se.64.768.128   & 0.688 & 0.067\\
se.48.768.128   & 0.646 & 0.069\\
se.32.1152.192  & 0.604 & 0.071\\
se.80.576.96    & 0.562 & 0.072\\
se.48.576.96    & 0.542 & 0.072\\
se.64.384.64    & 0.542 & 0.072\\
se.16.1344.224  & 0.542 & 0.072\\
se.32.960.160   & 0.521 & 0.072\\
se.16.1152.192  & 0.500 & 0.072\\
se.32.768.128   & 0.458 & 0.072\\
se.80.384.64    & 0.417 & 0.071\\
residual.40.256 & 0.375 & 0.070\\
se.32.576.96    & 0.375 & 0.070\\
se.48.384.64    & 0.375 & 0.070\\
se.16.960.160   & 0.354 & 0.069\\
se.32.384.64    & 0.292 & 0.066\\
se.16.768.128   & 0.271 & 0.064\\
se.16.576.96    & 0.229 & 0.061\\
residual.20.256 & 0.229 & 0.061\\
se.16.384.64    & 0.125 & 0.048\\
  \end{tabular}
\end{table}

The se.64.1152.192 network played games on the internet Kiseido Go Server (KGS) using only the policy to play and playing its moves instantly. Many people play against the network making it busy 24 hours a day. It reached a strong 6 dan level and it is still improving its rating, winning 80\% of its games. It is the best ranking we could reach with a policy network alone, the previous best ranking was 5 dan with a mobile network \cite{cazenave2020mobile}.

We now make the networks play using GPUCT and 32 descents per move. All networks play a round robin tournament. Table \ref{table32AgainstSE} gives the results for all networks. The se.64.1152.192 is again the best network and the residual networks are better than the smallest MobilNets but still far behind the best MobileNets.

\begin{table}
  \centering
  \caption{Making all networks play a round robin tournament with 32 descents at each move.}
  \label{table32AgainstSE}
  \begin{tabular}{lrrrrrrrrrr}
  
  Network                                    & Winrate & ~~~~~~~~$\frac{\sigma}{\sqrt{n}}$\\
                                             &        & \\
se.64.1152.192 & 0.812 & 0.056\\
se.48.960.160 & 0.812 & 0.056\\
se.48.1152.192 & 0.792 & 0.059\\
se.80.576.96 & 0.771 & 0.061\\
se.64.768.128 & 0.708 & 0.066\\
se.32.1344.224 & 0.667 & 0.068\\
se.32.1152.192 & 0.646 & 0.069\\
se.32.960.160 & 0.646 & 0.069\\
se.48.576.96 & 0.646 & 0.069\\
se.64.576.96 & 0.625 & 0.070\\
se.48.768.128 & 0.604 & 0.071\\
se.32.768.128 & 0.521 & 0.072\\
se.64.384.64 & 0.521 & 0.072\\
se.80.384.64 & 0.521 & 0.072\\
se.16.1152.192 & 0.396 & 0.071\\
se.16.960.160 & 0.375 & 0.070\\
se.16.768.128 & 0.333 & 0.068\\
se.32.576.96 & 0.333 & 0.068\\
se.48.384.64 & 0.333 & 0.068\\
se.16.1344.224 & 0.333 & 0.068\\
residual.40.256 & 0.271 & 0.064\\
residual.20.256 & 0.250 & 0.062\\
se.32.384.64 & 0.250 & 0.062\\
se.16.576.96 & 0.229 & 0.061\\
se.16.384.64 & 0.104 & 0.044\\
  \end{tabular}
\end{table}

In the last experiment we make all the networks use the same thinking time of 10 seconds per move on CPU. Large and slow networks make less descents than small and fast networks in this experiment. So there is a balance between the gain of accuracy of the policy and the improvement of the value due to increasing the size of the network and the slowdown due to the increased time for a forward pass. Results are given in Table \ref{table10AgainstSE}. Interestingly the se.64.1152.192 network is not the best network anymore. The residual networks are still way behind. We observe the impact of the balance between the size of the network and its speed.

\section{Conclusion}

We proposed a generalization of the PUCT bandit of AlphaGo and Alpha Zero so as to make it invariant to the number of descents. Experimental results show it is less sensitive to the budget than usual PUCT. We also proposed improvements to MobileNets and show that they give much better results than the commonly used residual networks. We made a detailed analysis of the balance between the depth, the width and the speed of MobileNets. We also made the networks play in order to evaluate their strengths.

In future work we plan to use MobileNets with Squeeze and Excitation in a Deep Reinforcement Learning framework, using the Expert Iteration algorithm to train the networks. We also plan to use Expert Iteration combined with GPUCT and MobileNets for various games and optimization problems.

\begin{table}
  \centering
  \caption{Making all networks play a round robin tournament with 10 seconds of CPU at each move.}
  \label{table10AgainstSE}
  \begin{tabular}{lrrrrrrrrrr}
  
  Network                                    &  Winrate & ~~~~~~~~~$\frac{\sigma}{\sqrt{n}}$\\
                                             &        &         & \\
se.48.768.128 & 0.688 & 0.067\\
se.64.384.64 & 0.646 & 0.069\\
se.64.576.96 & 0.625 & 0.070\\
se.32.768.128 & 0.604 & 0.071\\
se.48.960.160 & 0.604 & 0.071\\
se.80.576.96 & 0.604 & 0.071\\
se.32.1344.224 & 0.583 & 0.071\\
se.64.1152.192 & 0.583 & 0.071\\
se.32.1152.192 & 0.562 & 0.072\\
se.32.960.160 & 0.562 & 0.072\\
se.16.960.160 & 0.562 & 0.072\\
se.48.576.96 & 0.542 & 0.072\\
se.80.384.64 & 0.542 & 0.072\\
se.48.1152.192 & 0.521 & 0.072\\
se.32.384.64 & 0.521 & 0.072\\
se.16.1152.192 & 0.521 & 0.072\\
se.32.576.96 & 0.500 & 0.072\\
se.48.384.64 & 0.500 & 0.072\\
se.64.768.128 & 0.500 & 0.072\\
se.16.1344.224 & 0.500 & 0.072\\
se.16.768.128 & 0.479 & 0.072\\
se.16.576.96 & 0.333 & 0.068\\
se.16.384.64 & 0.229 & 0.061\\
residual.20.256 & 0.104 & 0.044\\
residual.40.256 & 0.083 & 0.040\\
  \end{tabular}
\end{table}



\bibliographystyle{IEEETran}
\bibliography{main}

\end{document}